\documentclass{article}

\usepackage[preprint]{neurips_2023}

\usepackage[utf8]{inputenc} 
\usepackage[T1]{fontenc}    
\usepackage{hyperref}       
\usepackage{url}            
\usepackage{booktabs}       
\usepackage{amsfonts}       
\usepackage{nicefrac}       
\usepackage{microtype}      
\usepackage[table]{xcolor}         
\usepackage{paralist}
\usepackage{color}
\usepackage{longtable}
\usepackage{graphicx}

\newcommand*{\scale}[2][4]{\scalebox{#1}{$#2$}}

\title{A Quantitative Review on \\ Language Model Efficiency Research}

\author{%
  Meng Jiang, Hy Dang, and Lingbo Tong \\
  Department of Computer Science and Engineering \\
  University of Notre Dame\\
  Notre Dame, IN 46556 \\
  \texttt{\{mjiang2, hdang, ltong2\}@nd.edu} \\
}

\begin{document}

\maketitle

\begin{abstract}
Language models (LMs) are being scaled and becoming powerful. Improving their efficiency is one of the core research topics in neural information processing systems. \citeauthor{tay2022efficient} provided a comprehensive overview of efficient Transformers that have become an indispensable staple in the field of NLP. However, in the section of \emph{On Evaluation}, they left an open question ``which fundamental efficient Transformer one should consider,'' answered by ``still a mystery,'' because ``many research papers select their own benchmarks.'' Unfortunately, there was not quantitative analysis about the performances of Transformers on any benchmarks. Moreover, state space models (SSMs) have demonstrated their abilities of modeling long-range sequences with non-attention mechanisms, which were not discussed in the prior review. This article makes a \emph{meta analysis} on the results from a set of papers on efficient Transformers as well as those on SSMs. It provides a quantitative review on LM efficiency research and gives suggestions for future research.

\end{abstract}

\section{Introduction}
Language models are trained to learn the underlying distribution of words in a given language.
A Transformer model is a neural network that learns context and thus meaning by tracking relationships in sequential data like the words in natural language~\citep{vaswani2017attention}. It has been observed that when the Transformers are scaled to have billions of parameters, the Transformer-based language models exhibit amazing performance on various language tasks~\citep{wei2022emergent}. Meanwhile, the efficiency of the language models, in terms of time and memory complexity, has attracted great attention from research, because it would address the bottleneck of training and deploying such large-scale models.
\cite{tay2022efficient} wrote a survey paper that provided a taxonomy of efficient Transformer models, characterizing them by their technical innovation and primary use case. The review was comprehensive, but unfortunately, readers who really wanted to learn or do \emph{language model efficiency research} would not be able to find answers to the following questions:
\begin{itemize}
    \item Q1: What were the state-of-the-art efficient language models?
    \item Q2: Were the results (i.e., performance measures) reported and confirmed by multiple sources? Were there \emph{inconsistent} results, i.e., significantly different measured performances of one type of solutions reported by different sources?
    \item Q3: Most studies claimed that they were the best solution when the papers were submitted or accepted. Were these claims correct?
\end{itemize}

That's because the solid answers would require a \emph{quantitative} analysis across papers in this research field. The existing survey, in \emph{On Evaluation} section on page 21, discussed a few NLP benchmarks such as GLUE, NaturalQuestions, and TriviaQA; and it argued that ``many research papers select their own benchmarks to showcase the abilities of the proposed model,'' leaving it ``a mystery to which fundamental efficient Transformer block one should consider using''~\citep{tay2022efficient}.

As dozens of related studies are being performed and published, some language tasks and datasets have been commonly selected as test scenarios to evaluate/compare model efficiency. Moreover, non-attention models such as state space models (SSMs)~\citep{gu2022efficiently} have been proposed to address the long range modeling problem and evaluated on the NLP benchmarks. Neither their technical innovations or experimental results were discussed in the existing survey. A broad, comprehensive, quantitative review is needed in language model efficiency research. And this article presents the work. It briefly describes different types of efficient language models, followed by tasks, datasets, and evaluation metrics. Then it provides a comprehensive quantitative meta analysis that aims to integrate the results from prior research to answer the aforementioned questions.

Reducing time and/or memory complexity would inevitably sacrifice a bit non-efficiency performance like accuracy. When the complexities of a set of models were reduced to a certain level,  one could claim that the model that achieved the highest accuracy would be the most efficient solution. It sacrificed the accuracy the least, so people hypothesize that when the models achieved the same accuracy, this model would have the least complexity. Past empirical studies unanimously performed efficiency evaluation based on this hypothesis. Our key observations from a meta analysis on these empirical results are listed as follows:
\begin{enumerate}
    \item Most empirical studies compared their proposed model against others on multiple tasks, and usually claimed theirs is the best one. However, the meta analysis identifies different winning approaches for different tasks and even different datasets. It fixes the one-sided understanding that researchers would have from learning only one or a few empirical studies.
    \item More than half of the results were reported by at least two sources. However, it is impossible to tell from the papers whether the numbers were reproduced/confirmed or just re-used from previous work. Meanwhile, inconsistent results were found on almost every task, caused by various settings of hyperparameters (e.g., model sizes, configurations) and reproductions.
    \item In quite a few studies, the proposed models were evaluated on a small subset of the tasks, and they actually did \emph{not} perform better than those that were published earlier and not cited.
\end{enumerate}

Based on this quantitative review, we offer a few suggestions for future research:
\begin{enumerate}
    \item Researchers are suggested to investigate as many suitable baselines as possible, write clearly if the numbers in experimental results were re-used from prior work or reproduced, and report and analyze any inconsistent results that are identified compared with related studies.
    \item Researchers in this field need a community and need a public collection of leaderboards.
    \item We hope that this quantitative review is helpful, and if it is, it should be continuously updated. Quantitative reviews are needed in flourishing research fields.
\end{enumerate}

\vspace{-0.02in}
\section{Efficient Language Model Architectures}
\vspace{-0.02in}
The core ability of language models is modeling and learning sequential dependencies which play an essential role in natural language. Besides traditional recurrent models (e.g., LSTM), there are two popular types of model architectures: Transformer models and state space models, corresponding to fully self-attention and non-attention mechanisms.

\vspace{-0.02in}
\subsection{Transformers}
\vspace{-0.01in}

The multi-headed self-attention in Transformers delivers improved parallelism, enabling more efficient processing of input sequences than recurrent models. \cite{tay2022efficient} provided a taxonomy of Transformer models in its Figure 2 and Section 3.1. The taxonomy has six categories based on core techniques that improve the memory complexity of the self-attention mechanism: (1) fixed/factorized patterns, e.g., Sparse Transformer~\citep{child2019generating}, (2) learnable patterns, e.g., Reformer~\citep{kitaev2020reformer}, (3) low rank/kernels, e.g., Linear Transformer~\citep{wang2020linformer}, (4) recurrence, e.g., Transformer-XL~\citep{dai2019transformer}, (5) memory/downsampling, e.g., Charformer~\citep{tay2022charformer}, and (6) sparse attention, e.g., Switch Transformer~\citep{fedus2022switch}.
Some Transformer models adopt multiple core techniques. Compressive Transformer~\citep{rae2019compressive} uses both memory/downsampling and recurrence to compress the Transformer parameters. Longformer~\citep{beltagy2020longformer} combines the pattern and downsampling approaches to reduce the memory complexity.

\begin{table}[t]
    \centering
    \includegraphics[width=0.97\textwidth]{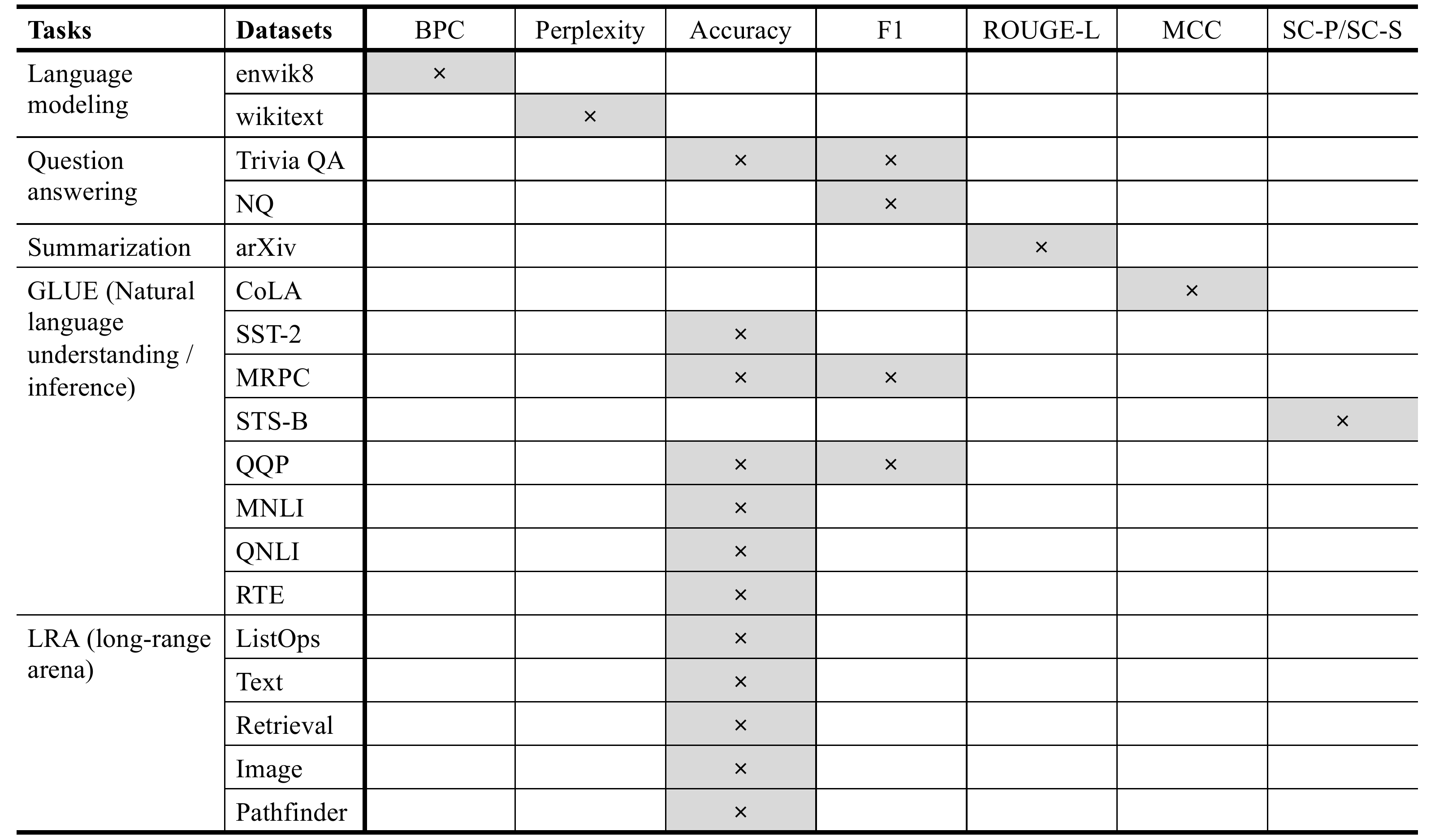}
    \caption{Most popular tasks, datasets, and evaluation metrics for language model efficiency research}
    \label{tab:metrics}
    \vspace{-0.2in}
\end{table}

\subsection{State Space Models}
\vspace{-0.1in}

State space models refer to a class of probabilistic graphical models that describes the probabilistic dependence between the latent state variable and the observed measurement. They deal with dynamic time series problems in control engineering and thus have the potential to deal with long-range dependency (LRD) problems in modeling natural language.
\cite{gu2021combining} introduce the Linear State-Space Layer (LSSL), which utilizes an implicit state to map a 1-dimensional input sequence to an output sequence through the simulation of a linear continuous-time state-space representation in discrete-time. LSSLs have been employed to simulate continuous processes, handle missing data, and adapt to different timescales. However, this work has been limited due to prohibitive computation and memory requirements induced by the state representation. This has led to the introduction of the Structured State Space sequence model (S4), which solves the critical computational bottleneck~\citep{gu2022efficiently}. The S4 model transforms the state matrices by breaking them down into a low-rank and normal terms. It computes truncated generating functions in frequency space, making it simpler to evaluate. S4 significantly advances the state-of-the-art for low-rank decomposition (LRD), long-range arena (LRA) tasks, and speech classification with long sequences. 

S4 achieves high performance, but
the diagonal-plus-low-rank structure
requires several reduction steps and linear algebraic techniques to compute state space output efficiently, making S4 difficult to analyze. A recent paper proposes a simpler Diagonal State Space (DSS) \citep{gupta2022diagonal} model that enforces diagonal state matrices, making it easier to formulate, implement, and analyze while being as expressive as general state spaces. \cite{hasani2022liquid} use a diagonal plus low-rank decomposition of the state transition matrix introduced in S4, and a few simplifications, the LTC-based structural state-space model, dubbed Liquid-S4, achieves the new state-of-the-art generalization across sequence modeling tasks with long-term dependencies such as text, image, and audio.

Recently, \cite{smith2022simplified} introduce a new state space layer -- the S5 layer -- which builds on the S4 layer but simplifies it in two main ways. First, S5 uses one multi-input multi-output SSM instead of the bank of many independent single-input single-output SSMs in S4. Second, S5 uses an efficient parallel scan instead of the convolutional and frequency-domain approach used by S4. The resulting S5 layer has the same computational complexity as S4 but operates purely recurrently and in the time domain. The final S5 layer has many desirable properties, including linear complexity in the sequence length, the ability to handle time-varying SSMs and irregularly sampled observations, and state-of-the-art performance on a variety of long-range sequence modeling tasks.

\begin{table}[t]
    \centering
    \caption{\textbf{Selected} results on \textbf{enwik8}, a language modeling task dataset. Smaller BPC (bits per character) is better. Bold numbers: claimed as the best results where they were proposed. Background colors: one color indicates one set of inconsistent results. Complete results are in Table~\ref{tab:full_enwik8}.}
    \label{tab:select_enwik8}
    \begin{tabular}{l|c|p{0.3\textwidth}}
    \toprule
    \textbf{Model} & \textbf{BPC $\downarrow$} & \textbf{Sources} \\
    \midrule
    LN HyperNetworks \citep{ha2016hypernetworks} & 1.34 & Table 2 in \citep{dai2019transformer}, Table 4 in \citep{rae2020compressive} \\
    \cellcolor{red!25}Locality-Sensitive Hashing \citep{kitaev2020reformer} & 1.33 & Table 5 in \citep{wang2021cluster} \\
    LN HM-LSTM \citep{chung2016hierarchical} & 1.32 & Table 2 in \citep{dai2019transformer}, Table 4 in \citep{rae2020compressive} \\
    RHN \citep{zilly2017recurrent} & 1.27 & Table 2 in \citep{dai2019transformer}, Table 4 in \citep{rae2020compressive} \\
    Large mLSTM / mLSTM \citep{krause2016multiplicative} & 1.24 & Table 2 in \citep{dai2019transformer} / Table 4 in \citep{rae2020compressive} \\
    Cluster-Former (\#C=512) \citep{wang2021cluster} & \textbf{1.22} & Table 5 in \citep{wang2021cluster} \\
    \cellcolor{yellow!25}T12 \citep{al2019character} & 1.11 & Table 4 in \citep{wang2021cluster} \\
    \cellcolor{yellow!25}64L Transformer / 64L Transf. \citep{al2019character} & 1.06 & Table 2 in \citep{dai2019transformer} / Table 4 in \citep{rae2020compressive} \\
    \cellcolor{blue!25}Transformer-XL / XFM-XL \citep{dai2019transformer} & 1.06 & Table 5 in \citep{wang2021cluster}, Table 5 in \citep{ma2023mega} \\
    \cellcolor{red!25}Reformer \citep{kitaev2020reformer} & 1.05 & Table 4 in \citep{zhu2021long} \\
    \cellcolor{green!25}Adaptive \citep{sukhbaatar2019adaptive} & 1.02 & Table 5 in \citep{wang2021cluster} \\
    MEGA \citep{ma2023mega} & \textbf{1.02} & Table 5 in \citep{ma2023mega} \\
    \cellcolor{blue!25}24L Transformer-XL / 24L TXL \citep{dai2019transformer} & \textbf{0.99} & Table 2 in \citep{dai2019transformer} / Table 4 in \citep{rae2020compressive} \\
    \cellcolor{green!25}Adaptive Transf. \citep{sukhbaatar2019adaptive} & 0.98 & Table 4 in \citep{rae2020compressive} \\
    24L Compressive Transformer \citep{rae2020compressive} & \textbf{0.97} & Table 4 in \citep{rae2020compressive} \\
    \bottomrule
    \end{tabular}
\end{table}

\begin{table}[ht]
    \centering
    \caption{\textbf{Selected} results of empirical studies on \textbf{NQ (long answer)}, a popular question answering dataset. Bigger F1 score is better. Results are reported on two evaluation sets: \textbf{Dev} and \textbf{Test}.}
    \label{tab:select_nq_long}
    \begin{tabular}{p{0.55\textwidth}|l|p{0.31\textwidth}}
    \toprule
    \textbf{Model} & \textbf{F1 $\uparrow$} & \textbf{Sources} \\ \midrule
    On \textbf{Dev}: & & \\
    DecAtt \citep{parikh2016decomposable} + DocReader \citep{chen2017reading} & 54.8 & Table 2 in \citep{zhang2021poolingformer}, Table 2 in \citep{wang2021cluster} \\
    BERT-large / BERT-large / BERT-joint \citep{alberti2019bert} & 64.7 & Table 2 in \citep{ainslie2020etc} / Table 2 in \citep{zhang2021poolingformer} / Table 2 in   \citep{wang2021cluster} \\
    Sparse Transformer / Sparse Attention \citep{jaszczur2021sparse} & 74.5 & Table 2 in  \citep{zhang2021poolingformer} / Table 2 in \citep{wang2021cluster} \\
    \cellcolor{red!25}{RikiNet-RoBERTa / RikiNet / RikiNet \citep{liu2020rikinet}} & 75.3 & Table 2 in \citep{wang2021cluster} / Table 2 in  \citep{ainslie2020etc} / Table 2 in \citep{zhang2021poolingformer} \\
    Reformer / Locality-Sensitive Hashing \citep{kitaev2020reformer} & 75.5 & Table 2 in   \citep{zhang2021poolingformer} / Table 2 in \citep{wang2021cluster} \\
    \cellcolor{red!25}{RikiNet-ensemble \citep{liu2020rikinet}} & \textbf{75.9} & Table 2 in   \citep{zhang2021poolingformer} \\
    Cluster-Former \citep{wang2021cluster} & \textbf{76.5} & Table 2 in \citep{zhang2021poolingformer}, Table 2 in \citep{wang2021cluster} \\
    ReflectionNet-ensemble \citep{wang2020no} & \textbf{77.0} & Table 2 in   \citep{zhang2021poolingformer} \\
    Poolingformer \citep{zhang2021poolingformer} & \textbf{77.5} & Table 2 in   \citep{zhang2021poolingformer} \\
    ETC-large (lifting from RoBERTa) \citep{ainslie2020etc} & \textbf{78.2} & Table 2 in \citep{ainslie2020etc} \\
    \midrule
    On \textbf{Test}: & & \\
    RikiNet-v2 / RikiNet-ensemble \citep{liu2020rikinet} & 76.1 & Table 3 in \citep{zaheer2020big} / Table 2 in \citep{zhang2021poolingformer} \\
    \cellcolor{blue!25}{ReflectionNet \citep{liu2020rikinet}} & 77.1 & Table 3 in \citep{zaheer2020big} \\
    \cellcolor{blue!25}{ReflectionNet-ensemble \citep{liu2020rikinet}} & 77.2 & Table 2 in \citep{zhang2021poolingformer} \\
    ETC (official) \citep{ainslie2020etc} & \textbf{77.78} & Table 5 in \citep{ainslie2020etc} \\
    BIGBIRD-ETC \citep{zaheer2020big} & \textbf{77.8} & Table 3 in \citep{zaheer2020big}, Table 2 in \citep{zhang2021poolingformer} \\
    Cluster-Former-ensemble / Cluster-Former \citep{wang2021cluster} & \textbf{78} & Table 3 in \citep{wang2021cluster} / Table 2 in \citep{zhang2021poolingformer} \\
    Poolingformer-ensamble \citep{zhang2021poolingformer} & \textbf{79.8} & Table 2 in \citep{zhang2021poolingformer} \\
    \bottomrule
    \end{tabular}
\end{table}

\begin{table}[t]
    \centering
    \caption{\textbf{Selected} results on \textbf{arXiv} for document summarization. Bigger R-L (ROUGE-L) is better.}
    \label{tab:select_arxiv}
    \begin{tabular}{p{0.52\textwidth}|c|p{0.33\textwidth}}
    \toprule
    \textbf{Model} & \textbf{R-L $\uparrow$} & \textbf{Sources} \\
    \midrule
    Long-Doc-Seq2Seq / Discourse-aware \citep{cohan2018discourse} & 31.80 & Table 4 in \citep{zaheer2020big} / Table 11 in \citep{beltagy2020longformer} \\
    Extr-Abst-TLM \citep{subramanian2019extractive} & 38.03 & Table 4 in \citep{zaheer2020big}, Table 11 in \citep{beltagy2020longformer}, Table 4 in \citep{zhang2021poolingformer} \\
    Sent-PTR\citep{subramanian2019extractive} & 38.06 & Table 4 in \citep{zaheer2020big}, Table 4 in \citep{zhang2021poolingformer} \\
    \cellcolor{red!25}Dancer / Dancer / Dancer RUM \citep{gidiotis2020divide} & 38.44 & Table 4 in \citep{zaheer2020big} / Table 4 in \citep{zhang2021poolingformer} / Table 5 in \citep{jaszczur2021sparse} \\
    \cellcolor{yellow!25}Pegasus \citep{zhang2020pegasus} & 38.83 & Table 4 in \citep{zaheer2020big}, Table 5 in \citep{jaszczur2021sparse}, Table 11 in \citep{beltagy2020longformer}, Table 4 in \citep{zhang2021poolingformer} \\
    \cellcolor{yellow!25}Pegasus (Re Eval) \citep{zhang2020pegasus} & 39.17 & Table 4 in \citep{zaheer2020big} \\
    \cellcolor{red!25}Dancer / Dancer PEGASUS \citep{gidiotis2020divide} & 40.56 & Table 4 in \citep{zhang2021poolingformer} / Table 5 in \citep{jaszczur2021sparse} \\
    BIGBIRD-Pegasus / BIGBIRD-Pegasus / BigBird (seqlen:4096) / BigBird \citep{zaheer2020big} &  \textbf{41.77} & Table 4 in \citep{zaheer2020big} / Table 5 in \citep{jaszczur2021sparse} / Table 11 in \citep{beltagy2020longformer} / Table 4 in \citep{zhang2021poolingformer} \\
    LED-large (seqlen: 16384) / LED16k \citep{beltagy2020longformer} & \textbf{41.83} & Table 11 in \citep{beltagy2020longformer} / Table 4 in \citep{zhang2021poolingformer} \\
    Poolingformer16k \citep{zhang2021poolingformer} & \textbf{42.69} & Table 4 in \citep{zhang2021poolingformer} \\
    \bottomrule
    \end{tabular}
\end{table}

\begin{table}[t]
    \centering
    \caption{\textbf{Complete} results on \textbf{GLUE QQP} (Quora Question Pairs2) for natural language understanding. Bigger (dev) Accuracy/F1 score is better. Results are sorted by dev F1.}
    \label{tab:full_glue_qqp}
    \begin{tabular}{p{0.48\textwidth}|c|c|p{0.31\textwidth}}
    \toprule
    \textbf{Model} & \textbf{Acc} & \textbf{F1} $\uparrow$ & \textbf{Sources} \\
    \midrule
    \cellcolor{red!25}BERT \citep{devlin2018bert} & 71.2 & - & Table 16 in \citep{zaheer2020big} \\
    Nystromformer \citep{xiong2021nystromformer} & -  & 86.3 & Table 2 in \citep{xiong2021nystromformer} \\
    \cellcolor{red!25}BERT-base\citep{devlin2018bert} & - & 87.3 & Table 2 in \citep{xiong2021nystromformer} \\
    DyConv\citep{wu2019pay} & 84.2 & 88.2 & Table 5 in \citep{Tay2020SynthesizerRS} \\
    BIGBIRD\citep{zaheer2020big} & 88.6 & - & Table 16 in \citep{zaheer2020big} \\
    \cellcolor{red!25}BERT-base (16GB) \citep{devlin2018bert} & - & 89.6 & Table 6 in \citep{ma2021luna} \\
    Linformer-128 (16GB) \citep{wang2020linformer} & - & 90.2 & Table 6 in \citep{ma2021luna} \\
    T5-Base+\citep{raffel2020exploring} & 88.3 & 91.2 & Table 5 in \citep{Tay2020SynthesizerRS} \\
    Luna-128 (160GB) \citep{ma2021luna} & - & 91.3 & Table 6 in \citep{ma2021luna} \\
    \cellcolor{yellow!25}XLNet \citep{yang2019xlnet} & 91.4 & - & Table 16 in \citep{zaheer2020big} \\
    Syn (D+V) \citep{Tay2020SynthesizerRS} & 88.6 & 91.5 & Table 5 in \citep{Tay2020SynthesizerRS} \\
    \cellcolor{blue!25}RoBERTa \citep{liu2019roberta} & 91.9 & - & Table 16 in \citep{zaheer2020big} \\
    \cellcolor{blue!25}ROBERTABase / RoBERTa-base (160GB) \citep{liu2019roberta} & - & 91.9 & Table 3 in \citep{dai2020funnel} / ~Table 6 in \citep{ma2021luna} \\
    Transformer (L24H1024) \citep{vaswani2017attention} & 89.6 & 92.2 & Table 2 in \citep{dai2020funnel} \\
    \cellcolor{blue!25}ROBERTALarge \citep{liu2019roberta} & - & 92.2 & Table 3 in \citep{dai2020funnel}\\
    \cellcolor{yellow!25}XLNetLarge \citep{yang2019xlnet} & - & 92.3 & Table 3 in \citep{dai2020funnel} \\
    ELECTRALarge \citep{clark2020electra} & - & 92.4 & Table 3 in \citep{dai2020funnel} \\
    Funnel Transformer (B10-10-10H1024) \citep{dai2020funnel} & \textbf{89.8} & \textbf{92.4} & Table 3 in \citep{dai2020funnel} \\
    \bottomrule
    \end{tabular}
\end{table}

\begin{table}[t]
    \centering
    \caption{\textbf{Selected} results on \textbf{LRA-Retrieval}. All inconsistent results and SSMs' results are selected.}
    \label{tab:select_lra_retrieval}
    \scale[0.88]{\begin{tabular}{p{0.45\textwidth}|l|p{0.52\textwidth}}
    \toprule
    \textbf{Model} & \textbf{Acc} $\uparrow$ & \textbf{Sources} \\
    \midrule
    \cellcolor{red!25}Linformer \citep{wang2020linformer} & 52.27 & Table 1 in \citep{ma2021luna}, Table 2 in \citep{ma2023mega}, Table 10 in \citep{gu2022efficiently} \\
    \cellcolor{red!25}Linformer \citep{wang2020linformer} & 53.09 & Table 1 in \citep{hasani2022liquid} \\
    \cellcolor{yellow!25}Reformer \citep{kitaev2020reformer} & 53.4 & Table 1 in \citep{hasani2022liquid}, Table 10 in \citep{gu2022efficiently}, Table 2 in \citep{ma2023mega}, Table 1 in \citep{ma2021luna} \\
    \cellcolor{blue!25}Performer in \citep{choromanski2020rethinking} & 53.82 & Table 1 in \citep{ma2021luna}, Table 2 in \citep{ma2023mega}, Table 10 in \citep{gu2022efficiently}, Table 1 in \citep{hasani2022liquid} \\
    \cellcolor{green!25}XFM / Transformer / Transformer / Transformer / Transformer \citep{vaswani2017attention} & 57.46 & Table 2 in \citep{ma2023mega} / Table 1 in \citep{hasani2022liquid} / Table 1 in \citep{ma2021luna} / Table 10 in \citep{gu2022efficiently} / Table 2 in \citep{smith2022simplified} \\
    \cellcolor{blue!25}Performer in \citep{choromanski2020rethinking} & 78.62 & Table 3 in \citep{xiong2021nystromformer}\\
    \cellcolor{yellow!25}Reformer \citep{kitaev2020reformer} & 78.64 & Table 1 in \citep{zhu2021long}, Table 3 in \citep{xiong2021nystromformer} \\
    \cellcolor{green!25}Transformer (re-impl) / XFM (re-impl) \citep{vaswani2017attention} & 79.14 & Table 1 in \citep{ma2021luna} / Table 2 in \citep{ma2023mega} \\
    \cellcolor{cyan!25}Luna-256 \citep{ma2021luna} & 79.29 & Table 10 in \citep{gu2022efficiently}, Table 2 in \citep{smith2022simplified}, Table 1 in \citep{hasani2022liquid} \\
    \cellcolor{green!25}Standard \citep{vaswani2017attention} & 79.35 & Table 3 in \citep{xiong2021nystromformer} \\ 
    \cellcolor{red!25}Linformer \citep{wang2020linformer} & 79.37 & Table 3 in \citep{xiong2021nystromformer}, Table 1 in \citep{zhu2021long} \\
    \cellcolor{cyan!25}Luna-256 \citep{ma2021luna} & \textbf{79.56} & Table 1 in \citep{ma2021luna}, Table 2 in \citep{ma2023mega} \\
    \cellcolor{gray!25}Nystromformer \citep{xiong2021nystromformer} & 79.56 & Table 3 in \citep{xiong2021nystromformer}, Table 1 in \citep{hasani2022liquid}, Table 10 in \citep{gu2022efficiently} \\
    \cellcolor{gray!25}Nystromformer \citep{xiong2021nystromformer} & 81.29 & Table 1 in \citep{zhu2021long} \\
    \cellcolor{blue!25}Performer in \citep{choromanski2020rethinking} & 81.7 & Table 1 in \citep{zhu2021long} \\
    \cellcolor{green!25}Full Attention \citep{vaswani2017attention} &82.3& Table 1 in \citep{zhu2021long} \\
    \cellcolor{orange!25}S4-v2 / S4 (updated) \citep{gu2022efficiently} & \textbf{90.9} & Table 2 in \citep{ma2023mega} / Table 10 in \citep{gu2022efficiently} \\   
    \cellcolor{orange!25}S4-v2 (re-impl) \citep{gu2022efficiently} & 90.94 & Table 2 in \citep{ma2023mega} \\
    Liquid-S4 / Liquid-S4-PB & 91.2 & Table 2 in \citep{smith2022simplified} / Table 1 in \citep{hasani2022liquid} \\
    MEGA \citep{ma2023mega} & \textbf{91.25} & Table 2 in \citep{ma2023mega}, Table 2 in \citep{smith2022simplified} \\
    S5 \citep{smith2022simplified} & \textbf{91.4} & Table 2 in \citep{smith2022simplified}, Table 1 in \citep{hasani2022liquid} \\
    \bottomrule
\end{tabular}}
\end{table}

\section{Meta Analysis}
\subsection{How to Read Our Results}

Table~\ref{tab:metrics} summarizes the NLP tasks that at least three studies have used to evaluate the efficiency of (proposed) language models, along with their typical datasets and evaluation metrics. Appendix~\ref{sec:appendixA} has more concrete descriptions. Results are presented from Table~\ref{tab:select_enwik8} to Table~\ref{tab:full_lra_pathfinder}. The results of the tables in this section are selected from those in Appendix, due to page limit. They have three columns: model ``mentions'', evaluation metric/score, and ``Sources'' for where the score was collected from, including the table number and paper citation. A score may be found in multiple sources. If the sources mention the model (with a citation in the first column) by different names, the names and sources are separated by ``/'' in the first and third columns, respectively. If the mentions are the same, the sources are separated by ``,'' instead.

We highlight a few things in the tables. First, the score is {bold} if the source claimed the proposed model achieved the best performance. So we can find answers to Q1 and Q3 in the Introduction. Second, some model mentions are highlighted with background colors such as \colorbox{red!25}{$\,$}, \colorbox{yellow!25}{$\,$}, \colorbox{blue!25}{$\,$}, \colorbox{green!25}{$\,$}, \colorbox{cyan!25}{$\,$}, \colorbox{gray!25}{$\,$}, and \colorbox{orange!25}{$\,$}, if there are inconsistent results (i.e., different evaluation scores) for the same model citation in different sets of sources. These extractions are helpful for Q2.

All the tables are sorted by evaluation scores, from the worst to the best. In the main sections, we select one dataset per task and select entries to make a table. An entry is selected if it satisfies any condition: (1) the score is bold for being claimed as the best; (2) the score is confirmed by more than one sources; (3) the model mentions are of background colors, associated with inconsistent results.

\vspace{-0.05in}
\subsection{Results on Language Modeling}

Table~\ref{tab:select_enwik8} contains results from five studies on the enwik8 dataset. \emph{Compressive Transformer} achieved the smallest BPC (0.97). Adaptive Transformer in~\citeyear{sukhbaatar2019adaptive} was the second best (0.98), better than Reformer in~\citeyear{kitaev2020reformer} (1.05), Cluster-Former in~\citeyear{wang2021cluster} (1.22), and MEGA in~\citeyear{ma2023mega} (1.02). There are four sets of highlighted results: although referencing to the same paper (which helps link the different model mentions), the results are different for the same dataset, e.g., Locality-Sensitive Hashing (1.33) and Reformer (1.05), Transformer-XL/XFM-XL (1.06) and 24L Transformer-XL/24L TXL (0.99), Adaptive (1.02) and Adaptive Transf. (0.98). But there was no discussion in any paper addressing the differences. The reason may be different mode configurations.

Table~\ref{tab:full_wikitext} in Appendix has results on the wikitext dataset. \emph{Routing Transformer} achieved the best PPL (15.8) but was not evaluated on enwik8. Compressive Transformer was ranked at the 2nd. As many as five different PPL scores were reported on the Adaptive Transformer model, citing the same paper~\citep{baevski2018adaptive}, which were worse than the PPL of proposed models in all the studies.







\subsection{Results on Question Answering}


Table~\ref{tab:select_nq_long} presents the results of various studies on {NQ (long answer)}. Among the models evaluated, ETC-large achieved the highest performance on the Dev set with an F1-score of 78.2 in \citeyear{ainslie2020etc}, followed by Poolingformer in \citeyear{zhang2021poolingformer} (77.5). Both models integrated neural memory modules, suggesting the potential of this approach. However, while Poolingformer (2021) claimed to be state-of-the-art in its original paper, it did not compare its performance with ETC-large (2020). On the Test set, Poolingformer-ensemble achieved the highest F1-score (79.8), followed by Cluster-Former (78.0), BIGBIRD-ETC (77.8), and ETC (77.78). The full meta results are in Table~\ref{tab:full_nq_long} in Appendix~\ref{sec:appendixB}.

Table~\ref{tab:full_nq_short} gives results from studies conducted on {NQ (short answer)}. ReflectionNet emerged as the best-performing model on both the Dev set (63.4) and the Test set (64.1), where it significantly outperformed all other models, including RikiNet (61.3) and Cluster-Former (60.9 on the Test set). The reason for the outstanding performance ReflectionNet is that it targeted particularly on the challenge of no-answer condition on the NQ dataset.

Upon comparing results on NQ long answer and short answer, it is evident that the best-performing models on the two sub tasks are different. For example, ETC and PoolingFormer outperformed all other models on NQ long answer, but was exceeded by ReflectionNet on NQ short answer. Similarly, ReflectionNet was the state-of-the-art method on NQ short answer but was not on NQ long answer. Interestingly, while the NQ long answer is generally considered more challenging than the short answer as it requires the model to understand the context of the question and return a longer piece of text, models generally achieve a higher F1 score on NQ long answer compared to NQ short answer.

Table~\ref{tab:full_triviaqa} presents the results of studies on {TriviaQA}. BIGBIRD achieved the best performance on all three sub-tasks with F1 scores of 79.5, 84.5, and 92.4 on Dev, Test, and Test Verified, respectively. The rankings on Test and Test Verified are consistent, with BIGBIRD-ETC being the best followed by Fusion-in-Decoder, SpanBERT, and Longformer. The scores on Test Verified are generally higher, indicating that identifying the absence of the answer in the contextual passage is still a challenge. 

Additionally, it was observed that Switch (published in \citeyear{fedus2022switch}), the most recent model in Table~\ref{tab:full_triviaqa}, only compared its accuracy with T5 in the original paper, making its comparison with other models unavailable. Furthermore, during the meta analysis, we discovered that approximately half of the papers on TriviaQA reported only accuracy, while the other half reported only F1 score. We recommend that researchers report both results to facilitate comparisons and enable a more comprehensive evaluation of the models.

To summarize, we found that memory-based models such as Poolingformer, ETC, and BIGBIRD showed the best performance on QA. In addition, there is a trend of model integration where models are combining multiple techniques to achieve better results. For example, Poolingformer incorporates both sliding window patterns and global memory modules, while BIGBIRD-ETC builds upon ETC and uses various techniques, including global memory, random attention, and local sliding windows.

One unique aspect of the QA task is its well-standardized nature in comparison to other tasks like GLUE. The QA task benefits from a clearly defined dev/test set split established by the original paper. In addition, papers introducing results on NQ and TriviaQA rarely contain unambiguous model names (the highlights in our QA tables indicates variations of the same model rather than conflicted results). This illustrates the importance of having a public benchmark for evaluation purposes. It is recommended that other NLP tasks follow a similar approach to ensure consistency and replicability.

\subsection{Results on Summarization}


Table~\ref{tab:select_arxiv} presents results from studies conducted on \textbf{arXiv}. Poolingformer16k achieved the highest ROUGE-L score (42.69), followed by LED-large (41.83) and BIGBIRD-Pegasus (41.77). Since Poolingformer is known as memory-based transformer, LED (or referred as Longformer) is a variation of sparse transformer, and BIGBIRD belongs to both categories, the results indicate the effectiveness of both memory-based and sparse-based approaches on this task. Full results can be found in Table~\ref{tab:full_arxiv}.

During the analysis, we observe that Table~\ref{tab:select_arxiv} reports varying results for the same model. For instance, the reported performance of Dancer differs across multiple papers, with scores ranging from 38.44 to 40.56. Similarly, there are discrepancies in the reported scores for Pegasus, with values of 38.83 and 39.17. These differences in performance may be due to variations in the dataset splits and/or hyperparameter settings used in the experiments.

\subsection{Results on GLUE}

Table \ref{tab:full_glue_qqp} presents the results obtained on the GLUE-QQP benchmark dataset. The Funnel Transformer with configuration (B10-10-10H1024) achieved the highest performance, with accuracy (Acc) and F1 score values of 89.8 and 92.4, respectively. Similarly, Funnel Transformers, which is a Down Sampling Transformer approach, yielded the best results among 5 out of 8 benchmark datasets, indicating the effectiveness of this approach. The results for the remaining GLUE benchmark datasets can be found in the Appendix section, specifically Table \ref{tab:full_glue_mnli} to Table \ref{tab:full_glue_mrpc}.

ELECTRALarge emerged as the second-best model on the GLUE benchmark datasets, demonstrating competitive performance compared to Funnel Transformers. ELECTRALarge secured the second position in 6 out of 8 GLUE benchmark datasets and achieved the state-of-the-art result on GLUE STS-B with a Pearson correlation coefficient (SC-P) of 92.6, as illustrated in Table \ref{tab:full_glue_stsb}.

It is worth noting that while multiple papers reference the same study, the reported results for the same dataset differ. For instance, different papers report varying scores for BERT-base on the GLUE QQP benchmark dataset, with values of 87.3 and 89.6 being mentioned.

Furthermore, not all proposed models evaluated their performance on all GLUE benchmark datasets. Among the studies reviewed, only two papers, focusing on Synthesizer and Funnel Transformer, respectively, assessed their results across most of the GLUE benchmark datasets, encompassing eight datasets for Synthesizer and nine datasets for Funnel Transformer. Additionally, during this investigation of the GLUE dataset, it was observed that despite claiming superior results compared to other baselines, Synthesizer did not compare its performance against earlier models such as XLNet or ROBERTA.

Considering the diverse characteristics of the GLUE datasets and the experimental settings for each model, direct comparisons of all achieved results pose challenges. For example, different models underwent pre-training on distinct datasets. Therefore, to enable more meaningful comparisons among models on the GLUE datasets, it is suggested to incorporate pre-training on the same datasets or include all recent models in the comparisons.

\subsection{Results on Long Range Arena}


Table \ref{tab:select_lra_retrieval} presents the results obtained from seven studies conducted on the LRA-Retrieval dataset. The remaining LRA benchmark datasets, namely ListOps, Text, Image, and Pathfinder, are provided in the Appendix. Specifically focusing on the LRA-Retrieval dataset, S5 achieved the highest performance with a score of 91.4, while MEGA secured the second-best for the same dataset with a score of 91.25.

The Appendix includes Table \ref{tab:full_lra_listops}, which showcases the results achieved on the LRA-ListOps dataset. MEGA attained the highest accuracy (63.14), followed by Liquid-S4/Liquid S4-PB as the second-best performer (62.75). Similarly, for the other benchmark datasets within the LRA category, including Text (Table \ref{tab:full_lra_text}), Image (Table \ref{tab:full_lra_image}), and Pathfinder (Table \ref{tab:full_lra_pathfinder}), MEGA achieved state-of-the-art results across all of them, with no reported conflicts in evaluations. These outcomes highlight the promising performance of the MEGA model within the LRA benchmark.

In addition to MEGA, the S4-variants and S4 models also demonstrated strong performance in LRA tasks, achieving competitive results compared to baselines. This underscores the robust performance of State Space Models.

Based on the results, the most recent models, namely MEGA and SSMs, exhibit strong performance on the LRAs-benchmark dataset. According to \citep{ma2023mega}, the multi-dimensional damped EMA utilized in MEGA can be considered a simplified variant of a state space model, establishing a close relationship between MEGA and S4. However, the difference lies in MEGA not relying on the HiPPO framework for parameter initialization, distinguishing it from S4 or S4D.

It is worth noting that despite referencing the same paper, different studies report varying results for the same dataset. For example, on the LRA-Retrieval benchmark dataset, Nystromformer's performance differs across different papers, with reported scores of 79.56 and 81.29. Similarly, the scores for XFM/Transformer (57.46), Transformer (re-impl) / XFM (re-impl) (79.14), Standard (79.35), and Full Attention (82.3) also exhibit discrepancies. However, the paper does not provide any discussion or analysis addressing these differences.

Despite these variations, the overall results indicate significant improvement over Transformer models by utilizing recent models such as SSMs and MEGA to capture long-range dependencies. These observations present alternative approaches for analyzing the long-range context.

\vspace{-0.03in}
\section{Discussions}
\vspace{-0.03in}
Key observations and suggestions have been summarized in the Introduction section. In this section, we discuss a possible action item to implement the suggestions and its limitations.

Indirect evaluation and comparison are not desired. The real objective should be time and space complexity when the focus is model efficiency. We should not compare their marginal differences on accuracy-based performance when the complexity of the models was not quantitatively reported or analyzed. The comparison would be meaningless when the model complexities were not comparable. Therefore, when resources and budget allow, we propose to organize a competition to have teams develop models for two or three selected tasks which do not have to cover all the tasks in this review. The competition organization will provide a platform so the model submissions will be performed and evaluated by the same hardware and software. We will specific a few complexity-based metrics to create leaderboads for the tasks. The submissions will be considered for the leaderboard, only when they achieve ``satisfactory performance'' on the corresponding task, which is defined as a threshold of accuracy/error-based metrics. The shortness of the direct evaluation and comparison is the financial cost of development of a great number of models, especially when there is very little room for the top models to optimize their time and space complexity.

\vspace{-0.03in}
\section{Conclusion}
\vspace{-0.03in}
This was the first literature review on language model efficiency research that used a quantitative method and meta analysis. It offered a set of integrated comparative results that would not be observed without such an effort. It covered both Transformer-based models and the emergent state space models. The meta analysis finally gave suggestions for future research.

\vspace{-0.05in}
\paragraph{Limitations} This quantitative review was developed upon the taxonomy-based literature review by \cite{tay2022efficient}. Per Transformer-based models, our review covers only the ones that have been studied and categorized in the previous review. We add a set of recent work on state space models. However, the paper collection was limited to April 1, 2023. We will continuously update this quantitative review when new methods or even new types of methods emerge.

\newpage

\bibliographystyle{plainnat}
\bibliography{ref}

\newpage

\section{Appendix: Evaluation Methods on Language Model Efficiency}
\label{sec:appendixA}
In this section, also as shown in Table~\ref{tab:metrics}, we summarize the NLP tasks that at least three studies have used to evaluate the efficiency of (proposed) language models, followed by the description of their typical datasets and evaluation metrics.

\subsection{Tasks}

\paragraph{Language Modeling} is a task that involves predicting the likelihood of a sequence of words in a language. The goal of language modeling is to learn the underlying patterns and structure of a language, which can be used to generate new text or to improve the performance of other NLP tasks.




\paragraph{Question Answering} involves answering a question posed in natural language. The goal is to build an automated system that can understand natural language questions and provide accurate answers.


\paragraph{Summarization} involves creating a condensed version of a longer piece of text while retaining its main ideas and important information. The goal is to make large amounts of information more accessible and easier to understand, while also saving time and effort.




\paragraph{Natural Language Understanding / Inference} provides the fundamental abilities in applications in machine translation, text categorization, speech recognition, and large-scale content analysis.

\paragraph{Long-Range Sequence Modeling} tests the abilities of learning dependencies in long-context scenarios, such as byte-level language modeling and byte-level document classification.

\subsection{Datasets}

\paragraph{enwik8}
consists of the first 100 million bytes of the English Wikipedia, which contains a diverse range of text, including articles, tables, and lists.
It includes a wide range of linguistic phenomena, including spelling and grammatical errors, proper names, technical jargon, and more.

\paragraph{wikitext}
contains over 100 million words from a larger and more diverse set of Wikipedia articles (named wikitext-103), processed for language modeling tasks.

\paragraph{TriviaQA}
consists of a large collection of trivia questions and their corresponding answers, sourced from a variety of sources, including quiz bowl competitions, trivia websites, and more.
Many of the questions require deep comprehension and reasoning skills, rather than simple
keyword matching.

\paragraph{NQ} 
consists of over 300,000 real, anonymized, naturally occurring questions and answers (NaturalQuestions), sampled from a wide range of sources, such as community forums, how-to sites, and news articles.
The questions are diverse and cover a wide range of topics,
from history to pop culture.

\paragraph{arXiv}
is a big collection of research papers in the fields of physics, mathematics, computer science, and other related disciplines, which is commonly used for document summarization.

\paragraph{GLUE}
short for General Language Understanding Evaluation, 
is a benchmark dataset
consisting nine different tasks that cover a range of natural language inference (NLI) or understanding (NLU) tasks such as sentiment analysis, natural language inference, paraphrase detection, and more:
\begin{itemize}
    \item \textbf{CoLA} (Corpus of Linguistic Acceptability) involves determining whether a given sentence is grammatically correct or not.
    \item \textbf{SST-2} (Stanford Sentiment Treebank) involves determining the sentiment (positive or negative) of a given sentence.
    \item \textbf{MRPC} (Microsoft Research Paraphrase Corpus) involves determining whether two given sentences are semantically equivalent or not.
    \item \textbf{STS-B} (Semantic Textual Similarity Benchmark) involves determining the degree of semantic similarity between two given sentences.
    \item \textbf{QQP} (Quora Question Pairs) involves determining whether two given questions are semantically equivalent or not.
    \item \textbf{MNLI} (Multi-Genre Natural Language Inference) involves determining the relationship between a given pair of sentences (entailment, contradiction, or neutral).
    \item \textbf{QNLI}  (Question NLI) involves determining the relationship between a given pair of sentences (entailment or not).
    \item \textbf{RTE} (Recognizing Textual Entailment) involves determining whether a given sentence entails another given sentence or not.
    \item \textbf{WNLI} (Winograd NLI) involves resolving pronoun references in a given sentence to determine whether it entails another given sentence or not.
\end{itemize}

\paragraph{LRA}
short for Long-Range Arena, is a benchmark dataset
consisting multiple tasks to evaluate sequence models on long-context scenarios:
\begin{itemize}
    \item \textbf{ListOps} is designed to investigate the abilities of modeling hierarchically structured data in a long-context scenario, specifically, the parsing ability of neural models.
    \item \textbf{Text} refers to byte-level text classification, dealing with compositionality as it is required to.
    \item \textbf{Retrieval} refers to byte-level document retrieval, learning compressed text representations.
    \item \textbf{Image} refers to image classification, requiring to learn the 2D spatial relations between input pixels, while presented as a 1D sequence of symbols.
    \item \textbf{Pathfinder} refers to learn long-range spatial dependencies, as a synthetic visual task motivated by cognitive psychology. We skip the extreme version (Path-X) and focus on text modeling tasks.
\end{itemize}


\subsection{Evaluation Metrics}

\paragraph{BPC}
short for Bits per Character, 
measures the average number of bits required to encode each character in a generated text sequence. It is used for evaluating the performance of language models on language modeling tasks. The lower the BPC score, the better the model is at generating text that closely matches the distribution of the training data.

\paragraph{Perplexity}
measures how well a probability model predicts a sample. In simpler terms, it measures how well a language model can predict the next word in a sequence given the previous words. A lower perplexity indicates that the language model is better at predicting the next word in a sequence.

\paragraph{Accuracy}
measures how many predictions of a language model are correct, which can be applied for a variety of tasks. So it's used in both GLUE and LRA benchmarks, as well as some other tasks.

\paragraph{F1} is a combined measure of precision and recall for binary classification. Question answering, retrieval, and sentiment prediction use F1 score for evaluation, for being considered as the classification tasks. The F1 score ranges from 0 to 1, with 1 being the best possible score.

\paragraph{ROUGE-L}
measures the overlap between the generated summary and the reference summary, where the reference summary is a human-written summary of the same document. Specifically, it measures the Longest common subsequence between the generated and reference summaries.

\paragraph{MCC}
stands for Matthews Correlation Coefficient, and it measures the correlation between the predicted and actual labels in a binary classification task. It is particularly useful when dealing with imbalanced datasets where one class is significantly more prevalent than the other.

\paragraph{SC-P/SC-S}
stands for Spearman's rank correlation coefficient on population or samples.
It
measures the strength of the association between two variables, ranging from -1 to 1.

\section{Appendix: Extra Results from Meta Analysis}
\label{sec:appendixB}
\paragraph{Results of empirical studies on \emph{language modeling}} Table~\ref{tab:full_enwik8} presents the BPC of 24 models by integrating results from four research papers on the \textbf{enwik8} dataset. Seven of the results are confirmed or re-used by more than one study. Important prior work that contributed a better results was ignored in some of the studies. And there are at least four sets of inconsistent results, i.e., different values for the same citation. Selected entries are discussed in main content of this paper.

\begin{table}[ht]
    \centering
    \caption{Results of empirical studies on \textbf{enwik8}, a dataset for language modeling task. Smaller BPC (bits per character) is better. Bold numbers: claimed as the best results where they were proposed. Background colors: one color indicates one set of inconsistent results.}
    \label{tab:full_enwik8}
    \begin{tabular}{l|c|p{0.3\textwidth}}
    \toprule
    \textbf{Model} & \textbf{BPC $\downarrow$} & \textbf{Sources} \\
    \midrule
    7L LSTM \citep{graves2013generating} & 1.67 & Table 4 in \citep{rae2020compressive} \\
    Sliding window & 1.34 & Table 5 in \citep{wang2021cluster} \\
    LN HyperNetworks \citep{ha2016hypernetworks} & 1.34 & Table 2 in \citep{dai2019transformer}, Table 4 in \citep{rae2020compressive} \\
    \cellcolor{red!25}Locality-Sensitive Hashing \citep{kitaev2020reformer} & 1.33 & Table 5 in \citep{wang2021cluster} \\
    LN HM-LSTM \citep{chung2016hierarchical} & 1.32 & Table 2 in \citep{dai2019transformer}, Table 4 in \citep{rae2020compressive} \\
    ByteNet \citep{kalchbrenner2016neural} & 1.31 & Table 4 in \citep{rae2020compressive} \\
    Sparse Attention \citep{child2019generating} & 1.29 & Table 5 in \citep{wang2021cluster} \\
    RHN \citep{zilly2017recurrent} & 1.27 & Table 2 in \citep{dai2019transformer}, Table 4 in \citep{rae2020compressive} \\
    FS-LSTM-4 \citep{mujika2017fast} & 1.25 & Table 2 in \citep{dai2019transformer} \\
    Large mLSTM / mLSTM \citep{krause2016multiplicative} & 1.24 & Table 2 in \citep{dai2019transformer} / Table 4 in \citep{rae2020compressive} \\
    cmix v13 \citep{knol2017cmix} & 1.23 & Table 2 in \citep{dai2019transformer} \\
    Cluster-Former (\#C=512) \citep{wang2021cluster} & \textbf{1.22} & Table 5 in \citep{wang2021cluster} \\
    \cellcolor{yellow!25}T12 \citep{al2019character} & 1.11 & Table 4 in \citep{wang2021cluster} \\
    \cellcolor{yellow!25}64L Transformer / 64L Transf. \citep{al2019character} & 1.06 & Table 2 in \citep{dai2019transformer} / Table 4 in \citep{rae2020compressive} \\
    \cellcolor{blue!25}Transformer-XL / XFM-XL \citep{dai2019transformer} & 1.06 & Table 5 in \citep{wang2021cluster}, Table 5 in \citep{ma2023mega} \\
    \cellcolor{red!25}Reformer \citep{kitaev2020reformer} & 1.05 & Table 4 in \citep{zhu2021long} \\
    \cellcolor{green!25}Adaptive \citep{sukhbaatar2019adaptive} & 1.02 & Table 5 in \citep{wang2021cluster} \\
    MEGA \citep{ma2023mega} & \textbf{1.02} & Table 5 in \citep{ma2023mega} \\
    BP-Transformer \citep{ye2019bp} & 1.02 & Table 5 in \citep{wang2021cluster} \\
    Longformer \citep{tay2021long} & 1.00 & Table 5 in \citep{wang2021cluster} \\
    \cellcolor{blue!25}24L Transformer-XL / 24L TXL \citep{dai2019transformer} & \textbf{0.99} & Table 2 in \citep{dai2019transformer} / Table 4 in \citep{rae2020compressive} \\
    \cellcolor{green!25}Adaptive Transf. \citep{sukhbaatar2019adaptive} & 0.98 & Table 4 in \citep{rae2020compressive} \\
    24L Compressive Transformer \citep{rae2020compressive} & \textbf{0.97} & Table 4 in \citep{rae2020compressive} \\
    \bottomrule
    \end{tabular}
\end{table}

Table~\ref{tab:full_wikitext} presents the perplexity of 28 models by integrating results from five research papers on the \textbf{wikitext} dataset. Nine of the results are confirmed or re-used by more than one study. This table also reflects missing important prior work and inconsistent results.

\paragraph{Results of empirical studies on \emph{document summarization}} Table~\ref{tab:full_arxiv} presents the ROUGE-L values of 19 models by integrating results from three research papers on the \textbf{arXiv} dataset. Eight of the results are confirmed or re-used by more than one study.

\begin{table}[ht]
    \centering
    \caption{Results of empirical studies on \textbf{wikitext}, a dataset for language modeling task. Smaller PPL (perplexity) is better. Bold numbers: claimed as the best results where they were proposed. Background colors: one color indicates one set of inconsistent results.}
    \label{tab:full_wikitext}
    \begin{tabular}{p{0.55\textwidth}|l|p{0.3\textwidth}}
    \toprule
    \textbf{Model} & \textbf{PPL $\downarrow$} & \textbf{Sources} \\
    \midrule
    LSTM & 48.7 & Table 5 in \citep{rae2020compressive}, Table 1 in \citep{dai2019transformer} \\
    Temporal CNN / TCN \citep{bai2018empirical} & 45.2 & Table 5 in \citep{rae2020compressive} / Table 1 in \citep{dai2019transformer} \\
    LSTMs / LSTM + Neural cache \citep{grave2016improving} & 40.8 & Table 2 in \citep{roy2021efficient} / Table 1 in \citep{dai2019transformer} \\
    GLU CNN / GCNN-14 / GCNN-14 \citep{dauphin2017language} & 37.2 & Table 8 in \citep{gu2022efficiently} / Table 5 in \citep{rae2020compressive} / Table 1 in \citep{dai2019transformer} \\
    AWD-QRNN / Quasi-RNN / QRNNs / QRNN \citep{bradbury2016quasi} & 33 & Table 8 in \citep{gu2022efficiently} / Table 5 in \citep{rae2020compressive} / Table 2 in \citep{roy2021efficient} / Table 1 in \citep{dai2019transformer} \\
    RMC \citep{santoro2018relational} & 31.9 & Table 5 in \citep{rae2020compressive} \\
    \cellcolor{red!25}Hebbian + Cache \citep{rae2018fast} & 29.9 & Table 1 in \citep{dai2019transformer} \\
    \cellcolor{red!25}LSTM + Hebb. \citep{rae2018fast} & 29.2 & Table 8 in \citep{gu2022efficiently}, Table 5 in \citep{rae2020compressive} \\
    TrellisNet \citep{bai2018trellis} & 29.19 & Table 8 in \citep{gu2022efficiently} \\
    \cellcolor{yellow!25}Transformer \citep{baevski2018adaptive} & 26.2 & Table 1 in \citep{peng2021random} \\
    Dynamic Conv \citep{wu2019pay} & 25 & Table 8 in \citep{gu2022efficiently} \\
    RFA \citep{peng2021random} & \textbf{23.5} & Table 1 in \citep{peng2021random} \\
    TaLK Conv \citep{lioutas2020time} & 23.3 & Table 8 in \citep{gu2022efficiently} \\
    S4 \citep{gu2022efficiently} & 21.28 & Table 8 in \citep{gu2022efficiently}, Table 5 in \citep{ma2023mega}  \\
    Sliding window & 20.8 & Table 5 in \citep{wang2021cluster} \\
    Locality-Sensitive Hashing \citep{kitaev2020reformer} & 20.8 & Table 5 in \citep{wang2021cluster} \\
    Adaptive Transformer \citep{sukhbaatar2019adaptive} & 20.6 & Table 2 in \citep{roy2021efficient} \\
    \cellcolor{yellow!25}Transformer \citep{baevski2018adaptive} & 20.51 & Table 8 in \citep{gu2022efficiently} \\
    Sparse Attention \citep{child2019generating} & 20.5 & Table 5 in \citep{wang2021cluster}\\
    \cellcolor{yellow!25}Adaptive Input \citep{baevski2018adaptive} & 20.5 & Table 1 in \citep{dai2019transformer} \\
    Clusterformer \citep{wang2021cluster} & \textbf{20.2} & Table 4 in \citep{wang2021cluster} \\
    Local Transformer \citep{vaswani2017attention} & 19.8 & Table 2 in \citep{roy2021efficient} \\
    \cellcolor{yellow!25}Transformer / Adaptive Input \citep{baevski2018adaptive} & 18.7 & Table 5 in \citep{rae2020compressive} / Table 2 in \citep{roy2021efficient}\\
    \cellcolor{yellow!25}XFM-adaptive \citep{baevski2018adaptive,al2019character} & 18.66 & Table 5 in \citep{ma2023mega} \\
    XFM-XL / 18L TransformerXL / TransformerXL / TransformerXL Large \citep{dai2019transformer} & \textbf{18.3} & Table 5 in \citep{ma2023mega} / Table 5 in \citep{rae2020compressive} / Table 2 in \citep{roy2021efficient} / Table 1 in \citep{dai2019transformer} \\
    MEGA \citep{ma2023mega} & \textbf{18.07} & Table 5 in \citep{ma2023mega} \\
    Compressive Transformer \citep{rae2020compressive} & \textbf{17.1} & Table 5 in \citep{rae2020compressive} \\
    Routing Transformer \citep{roy2021efficient} & \textbf{15.8} & Table 2 in \citep{roy2021efficient} \\
    \bottomrule
    \end{tabular}
\end{table}

\begin{table}[ht]
    \centering
    \caption{Results of empirical studies on \textbf{arXiv}, a dataset for document summarization task. Bigger R-L (ROUGE-L) is better. Bold numbers: claimed as the best results where they were proposed. Background colors: one color indicates one set of inconsistent results.}
    \label{tab:full_arxiv}
    \begin{tabular}{p{0.52\textwidth}|c|p{0.33\textwidth}}
    \toprule
    \textbf{Model} & \textbf{R-L $\uparrow$} & \textbf{Sources} \\
    \midrule
    Pntr-Gen-Seq2Seq \citep{see2017get} & 25.16 & Table 4 in \citep{zaheer2020big} \\
    Attn-Seq2Seq \citep{sutskever2014sequence} & 25.56 & Table 4 in \citep{zaheer2020big} \\
    Transformer \citep{vaswani2017attention} & 25.58 & Table 4 in \citep{zaheer2020big} \\
    LSA \citep{wiseman2017challenges} & 25.67 & Table 4 in \citep{zaheer2020big} \\
    SumBasic \citep{nenkova2005impact} & 26.30 & Table 4 in \citep{zaheer2020big} \\
    LexRank \citep{erkan2004lexrank} & 28.99 & Table 4 in \citep{zaheer2020big} \\
    Transformer + RoBERTa\citep{rothe2020leveraging} & 29.53 & Table 4 in \citep{zaheer2020big} \\
    Transformer + Pegasus \citep{zhang2020pegasus} & 30.14 & Table 4 in \citep{zaheer2020big} \\
    Long-Doc-Seq2Seq / Discourse-aware \citep{cohan2018discourse} & 31.80 & Table 4 in \citep{zaheer2020big} / Table 11 in \citep{beltagy2020longformer} \\
    Extr-Abst-TLM \citep{subramanian2019extractive} & 38.03 & Table 4 in \citep{zaheer2020big}, Table 11 in \citep{beltagy2020longformer}, Table 4 in \citep{zhang2021poolingformer} \\
    Sent-PTR\citep{subramanian2019extractive} & 38.06 & Table 4 in \citep{zaheer2020big}, Table 4 in \citep{zhang2021poolingformer} \\
    \cellcolor{red!25}Dancer / Dancer / Dancer RUM \citep{gidiotis2020divide} & 38.44 & Table 4 in \citep{zaheer2020big} / Table 4 in \citep{zhang2021poolingformer} / Table 5 in \citep{jaszczur2021sparse} \\
    \cellcolor{yellow!25}Pegasus \citep{zhang2020pegasus} & 38.83 & Table 4 in \citep{zaheer2020big}, Table 5 in \citep{jaszczur2021sparse}, Table 11 in \citep{beltagy2020longformer}, Table 4 in \citep{zhang2021poolingformer} \\
    \cellcolor{yellow!25}Pegasus (Re Eval) \citep{zhang2020pegasus} & 39.17 & Table 4 in \citep{zaheer2020big} \\
    \cellcolor{red!25}Dancer / Dancer PEGASUS \citep{gidiotis2020divide} & 40.56 & Table 4 in \citep{zhang2021poolingformer} / Table 5 in \citep{jaszczur2021sparse} \\
    Terraformer \citep{jaszczur2021sparse} & 41.21 & Table 5 in \citep{jaszczur2021sparse} \\
    BIGBIRD-Pegasus / BIGBIRD-Pegasus / BigBird (seqlen:4096) / BigBird \citep{zaheer2020big} &  \textbf{41.77} & Table 4 in \citep{zaheer2020big} / Table 5 in \citep{jaszczur2021sparse} / Table 11 in \citep{beltagy2020longformer} / Table 4 in \citep{zhang2021poolingformer} \\
    LED-large (seqlen: 16384) / LED16k \citep{beltagy2020longformer} & \textbf{41.83} & Table 11 in \citep{beltagy2020longformer} / Table 4 in \citep{zhang2021poolingformer} \\
    Poolingformer16k \citep{zhang2021poolingformer} & \textbf{42.69} & Table 4 in \citep{zhang2021poolingformer} \\
    \bottomrule
    \end{tabular}
\end{table}

\paragraph{Results of empirical studies on \emph{natural language understanding / inference}} Table~\ref{tab:full_glue_qqp}, Table~\ref{tab:full_glue_mnli}, Table~\ref{tab:full_glue_sst2}, Table~\ref{tab:full_glue_qnli}, Table~\ref{tab:full_glue_cola}, Table~\ref{tab:full_glue_rte}, Table~\ref{tab:full_glue_stsb}, and Table~\ref{tab:full_glue_mrpc} present the results on eight natural language understanding and/or inference tasks in the \textbf{GLUE} benchmark:
\begin{compactitem}
\item \textbf{QQP}: Quora Question Pairs2, evaluated by dev accuracy and F1 score;
\item \textbf{MNLI}: Multi-Genre Natural Language Inference, evaluated by accuracy on match (m) and mismatch (mm);
\item \textbf{SST-2}: Stanford Sentiment Treebank, about sentiment analysis, evaluated by accuracy;
\item \textbf{QNLI}: Question-answering Natural Language Inference, evaluated by dev accuracy;
\item \textbf{COLA}: Corpus of Linguistic Acceptability, evaluated by dev Matthew's correlation coefficient (MCC);
\item \textbf{RTE}: Recognizing Textual Entailment, evaluated by accuracy;
\item \textbf{STS-B}: (Semantic Textual Similarity Benchmark, evaluated by dev Spearman correlation P (SC-P) and dev Spearman correlation S (SC-S);
\item \textbf{MRPC}: Microsoft Research Paraphrase Corpus, evaluated by dev accuracy and F1 score.
\end{compactitem}

\paragraph{Results of empirical studies on \emph{question answering}} There are two datasets that are commonly used to evaluate the efficiency of language models. TriviaQA is a challenging reading comprehension dataset containing over 650K question-answer-evidence triples, which was collected by University of Washington and published in 2017~\cite{joshi2017triviaqa}. The other is Natural Questions (NQ), a benchmark for question answering research collected by Google and published in 2019~\cite{kwiatkowski2019natural}.

Table~\ref{tab:full_triviaqa} presents the quantitative results on the \textbf{TriviaQA} dataset. The evaluation metrics are accuracy and F1 score. Different studies may use different metrics. Towards evaluation, the studies may use the development/validation set (Dev), test set (Test), or verified test set (Test Verified). And the results across different sets are not comparable.

Table~\ref{tab:full_nq_long} and Table~\ref{tab:full_nq_short} present the quantitative results on the \textbf{NQ} datasets, \textbf{long answers} and \textbf{short answers}, respectively. The studies use the Dev and Test sets for evaluation. The studies were not as many as those on other tasks, and no inconsistent results were observed.

\paragraph{Results of empirical studies on \emph{long-range arena}} Table~\ref{tab:full_lra_listops}, Table~\ref{tab:full_lra_text}, Table~\ref{tab:full_lra_retrieval}, Table~\ref{tab:full_lra_image}, and Table~\ref{tab:full_lra_pathfinder} present the results on five tasks in the \textbf{LRA} benchmark, such as \textbf{ListOps}, \textbf{Text}, \textbf{Retrieval}, \textbf{Image}, and \textbf{Pathfinder}, all evaluated by accuracy.
Table \ref{tab:select_lra_retrieval} has results on the LRA-Retrieval dataset, which focuses on 

\begin{table}[ht]
    \centering
    \caption{Results of empirical studies on \textbf{GLUE MNLI} (Multi-Genre Natural Language Inference). Bigger (dev) Accuracy on match (m) / Accuracy on mismatch (mm) is better.}
    \label{tab:full_glue_mnli}
    \scale[0.86]{
\begin{tabular}{p{0.48\textwidth}|l|l|p{0.31\textwidth}}
    \toprule
    \textbf{Model} & \textbf{Acc (m)} $\uparrow$  & \textbf{Acc (mm)}  & \textbf{Sources} \\
    \midrule
    DyConv \citep{wu2019pay} & 73.8 & 75.1 & Table 5 in \citep{Tay2020SynthesizerRS} \\
    Nystromformer \citep{xiong2021nystromformer} & 80.9 & 82.2 & Table 2 in \citep{xiong2021nystromformer} \\
    \cellcolor{red!25}BERT-base\citep{devlin2018bert} & 82.4 & 82.4 & Table 2 in \citep{xiong2021nystromformer} \\
    \cellcolor{red!25}BERT \citep{devlin2018bert} & 84.6 & 83.4 & Table 16 in \citep{zaheer2020big} \\
    T5-Base \citep{raffel2020exploring} & 84.7 & 85 & Table 5 in \citep{Tay2020SynthesizerRS} \\
    Syn (R+V) \citep{Tay2020SynthesizerRS} & \textbf{85} & \textbf{84.6} & Table 5 in \citep{Tay2020SynthesizerRS} \\
    \cellcolor{yellow!25} XLNet \citep{yang2019xlnet} & 86.8 &  - & Table 16 in \citep{zaheer2020big} \\
    BIGBIRD \citep{zaheer2020big} & 87.5 & 87.3 & Table 16 in \citep{zaheer2020big} \\
    \cellcolor{blue!25}ROBERTABase / RoBERTa \citep{liu2019roberta} & 87.6 & - & Table 3 in \citep{dai2020funnel} / Table 16 in \citep{zaheer2020big} \\
    MPNetBase \citep{song2020mpnet} & 88.5 & - & Table 3 in \citep{dai2020funnel} \\
    Transformer \citep{vaswani2017attention} & 89.4 & - & Table 2 in \citep{dai2020funnel} \\
    \cellcolor{blue!25}ROBERTALarge \citep{liu2019roberta} & 90.2 & - & Table 3 in \citep{dai2020funnel}\\
    \cellcolor{yellow!25}XLNetLarge \citep{yang2019xlnet} & 90.8 & - & Table 3 in \citep{dai2020funnel} \\
    ELECTRALarge \citep{clark2020electra} & 90.9 & - & Table 3 in \citep{dai2020funnel} \\
    Funnel Transformer (B10-10-10H1024) \citep{dai2020funnel} & \textbf{91.1} & - & Table 3 in \citep{dai2020funnel} \\
    \bottomrule
    \end{tabular}}
\end{table}

\begin{table}[ht]
    \centering
    \caption{Results of empirical studies on \textbf{GLUE SST-2} (Stanford Sentiment Treebank), a dataset for sentiment analysis. Bigger accuracy is better.}
    \label{tab:full_glue_sst2}
    \scale[0.86]{\begin{tabular}{p{0.66\textwidth}|l|p{0.31\textwidth}}
    \toprule
    \textbf{Model} & \textbf{Acc} $\uparrow$ & \textbf{Sources} \\
    \midrule
    \cellcolor{red!25}BERT-base \citep{devlin2018bert} & 90 & Table 2 in \citep{xiong2021nystromformer} \\
    DyConv \citep{wu2019pay} & 90.6 & Table 5 in \citep{Tay2020SynthesizerRS} \\
    Nystromformer \citep{xiong2021nystromformer} & \textbf{91.4} & Table 2 in \citep{xiong2021nystromformer} \\
    Syn (D+V) \citep{Tay2020SynthesizerRS} & \textbf{92.4} & Table 5 in \citep{Tay2020SynthesizerRS} \\
    Linformer-128 (16GB) \citep{wang2020linformer} & 92.4 & Table 6 in \citep{ma2021luna} \\
    \cellcolor{red!25}BERT-base \citep{devlin2018bert} & 92.7 & Table 6 in \citep{ma2021luna} \\
    T5-Base+\citep{raffel2020exploring} & 92.9 & Table 5 in \citep{Tay2020SynthesizerRS} \\
\bottomrule
    \end{tabular}}
\end{table}

\begin{table}[ht]
    \centering
    \caption{Results of empirical studies on \textbf{GLUE COLA} (Corpus of Linguistic Acceptability). Bigger (dev) Matthew's correlation coefficient (MCC) is better.}
    \label{tab:full_glue_cola}
    \scale[0.86]{\begin{tabular}{p{0.66\textwidth}|l|p{0.31\textwidth}}
    \toprule
    \textbf{Model} & \textbf{MCC} $\uparrow$ & \textbf{Sources} \\
    \midrule
    DyConv \citep{wu2019pay} & 33.9 & Table 5 in \citep{Tay2020SynthesizerRS} \\
    BERT \citep{devlin2018bert} & 52.1 & Table 16 in \citep{zaheer2020big} \\
    Syn (R+V) \citep{Tay2020SynthesizerRS} & \textbf{53.3} & Table 5 in \citep{Tay2020SynthesizerRS} \\
    T5-Base+ \citep{raffel2020exploring} & 54.3 & Table 5 in \citep{Tay2020SynthesizerRS} \\
    BIGBIRD \citep{zaheer2020big} & 58.5 & Table 16 in \citep{zaheer2020big} \\
    \cellcolor{red!25}XLNet \citep{yang2019xlnet} & 60.2 & Table 16 in \citep{zaheer2020big} \\
    \cellcolor{yellow!25}ROBERTABase / RoBERTa \citep{liu2019roberta} & 63.6 & Table 3 in \citep{dai2020funnel} / Table 16 in \citep{zaheer2020big} \\
\bottomrule
    \end{tabular}}
\end{table}

\begin{table}[ht]
    \centering
    \caption{Results of empirical studies on \textbf{GLUE QNLI} (Question-answering Natural Language Inference). Bigger (dev) accuracy is better.}
    \label{tab:full_glue_qnli}
    \scale[0.9]{\begin{tabular}{p{0.62\textwidth}|l|p{0.31\textwidth}}
    \toprule
    \textbf{Model} & \textbf{Acc} $\uparrow$ & \textbf{Sources} \\
    \midrule
    DyConv \citep{wu2019pay} & 84.4 & Table 5 in \citep{Tay2020SynthesizerRS} \\
    \cellcolor{red!25}BERT-base \citep{devlin2018bert} & 88.4 & Table 6 in \citep{ma2021luna} \\
    Nystromformer \citep{xiong2021nystromformer} & 88.7 & Table 2 in \citep{xiong2021nystromformer} \\
    \cellcolor{red!25}BERT-base \citep{devlin2018bert} & 90.3 & Table 2 in \citep{xiong2021nystromformer} \\
    Linformer-128 (16GB) \citep{wang2020linformer} & 90.4 & Table 6 in \citep{ma2021luna} \\
    \cellcolor{red!25}BERT \citep{devlin2018bert} & 90.5 & Table 16 in \citep{zaheer2020big} \\
    T5-Base \citep{raffel2020exploring} & 91.7 & Table 5 in \citep{Tay2020SynthesizerRS} \\
\bottomrule
    \end{tabular}}
\end{table}

\begin{table}[ht]
    \centering
    \caption{Results of empirical studies on \textbf{GLUE RTE} (Recognizing Textual Entailment).}
    \label{tab:full_glue_rte}
    \scale[0.9]{\begin{tabular}{p{0.62\textwidth}|l|p{0.31\textwidth}}
    \toprule
    \textbf{Model} & \textbf{Acc} $\uparrow$ & \textbf{Sources} \\
    \midrule
    DyConv \citep{wu2019pay} & 58.1 & Table 5 in \citep{Tay2020SynthesizerRS} \\
    BERT \citep{devlin2018bert} & 66.4 & Table 16 in \citep{zaheer2020big} \\
    \cellcolor{red!25}XLNet \citep{yang2019xlnet} & 74 & Table 16 in \citep{zaheer2020big} \\
    BIGBIRD \citep{zaheer2020big} & 75 & Table 16 in \citep{zaheer2020big} \\
    \cellcolor{yellow!25}ROBERTABase / RoBERTa \citep{liu2019roberta} & 78.7 & Table 3 in \citep{dai2020funnel} / Table 16 in \citep{zaheer2020big} \\
    T5-Base+ \citep{raffel2020exploring} & 79.1 & Table 5 in \citep{Tay2020SynthesizerRS} \\
    Syn (R+V) \citep{Tay2020SynthesizerRS} & \textbf{81.2} & Table 5 in \citep{Tay2020SynthesizerRS} \\
    Transformer (L24H1024) \citep{vaswani2017attention} & 84.5 & Table 2 in \citep{dai2020funnel} \\
    MPNetBase \citep{song2020mpnet} & 85.2 & Table 3 in \citep{dai2020funnel} \\
    \cellcolor{red!25}XLNetLarge \citep{yang2019xlnet} & 85.9 & Table 3 in \citep{dai2020funnel} \\
    \cellcolor{yellow!25}ROBERTALarge \citep{liu2019roberta} & 86.6 & Table 3 in \citep{dai2020funnel} \\
    ELECTRALarge \citep{clark2020electra} & 88 & Table 3 in \citep{dai2020funnel} \\
    Funnel Transformer (B10-10-10H1024) \citep{dai2020funnel} & \textbf{89.5} & Table 3 in \citep{dai2020funnel} \\
    \bottomrule
    \end{tabular}}
\end{table}

\begin{table}[ht]
    \centering
    \caption{Results of empirical studies on \textbf{GLUE STS-B} (Semantic Textual Similarity Benchmark). Bigger (dev) Spearman correlation P (SC-P) / Spearman correlation S (SC-S) is better.}
    \label{tab:full_glue_stsb}
    \scale[0.9]{\begin{tabular}{p{0.52\textwidth}|l|l|p{0.31\textwidth}}
    \toprule
    \textbf{Model} & \textbf{SC-P} $\uparrow$ &  \textbf{SC-S} &  \textbf{Sources} \\
    \midrule
    DyConv \citep{wu2019pay} & 60.7 & 63.1 & Table 5 in \citep{Tay2020SynthesizerRS} \\
    BERT \citep{devlin2018bert} & 85.8 & - & Table 16 in \citep{zaheer2020big} \\
    BIGBIRD \citep{zaheer2020big} & 87.8 & - & Table 16 in \citep{zaheer2020big} \\
    T5-Base \citep{raffel2020exploring} & 89.1 & 88.9 & Table 5 in \citep{Tay2020SynthesizerRS} \\
    Syn (R+V) \citep{Tay2020SynthesizerRS} & \textbf{89.3} & \textbf{88.9} & Table 5 in \citep{Tay2020SynthesizerRS} \\
    \cellcolor{red!25}XLNet \citep{yang2019xlnet} & 89.5 & - & Table 16 in \citep{zaheer2020big} \\
    MPNetBase \citep{song2020mpnet} & 90.9 & - & Table 3 in \citep{dai2020funnel} \\
    \cellcolor{yellow!25}ROBERTABase / RoBERTa \citep{liu2019roberta} & 91.2 & - & Table 3 in \citep{dai2020funnel} / Table 16 in \citep{zaheer2020big} \\
    Transformer (L24H1024) \citep{vaswani2017attention} & 91.5 & - & Table 2 in \citep{dai2020funnel} \\
    Funnel Transformer (B10-10-10H1024) \citep{dai2020funnel} & 92.1 & - & Table 3 in \citep{dai2020funnel}  \\
    \cellcolor{yellow!25}ROBERTALarge \citep{liu2019roberta} & 92.4 & - & Table 3 in \citep{dai2020funnel} \\
    \cellcolor{red!25}XLNetLarge \citep{yang2019xlnet} & 92.5 & - & Table 3 in \citep{dai2020funnel} \\
    ELECTRALarge \citep{clark2020electra} & 92.6 & - & Table 3 in \citep{dai2020funnel} \\
    \bottomrule
    \end{tabular}}
\end{table}

\begin{table}[ht]
    \centering
    \caption{Results of empirical studies on \textbf{GLUE MRPC} (Microsoft Research Paraphrase Corpus). Bigger (dev) accuracy/F1 is better. The results are sorted by F1 score.}
    \label{tab:full_glue_mrpc}
    \begin{tabular}{p{0.5\textwidth}|l|l|p{0.31\textwidth}}
    \toprule
    \textbf{Model} & \textbf{Acc} & \textbf{F1} $\uparrow$ & \textbf{Sources} \\
    \midrule
    DyConv\citep{wu2019pay} & 72.5 & 75.1 & Table 5 in \citep{Tay2020SynthesizerRS} \\
    Syn (D+V) \citep{Tay2020SynthesizerRS} & 87.7 & 84.8 & Table 5 in \citep{Tay2020SynthesizerRS} \\
    T5-Base \citep{raffel2020exploring} & 88.7 & 85 & Table 5 in \citep{Tay2020SynthesizerRS} \\
    Nystromformer \citep{xiong2021nystromformer} & - & 88.1 & Table 2 in \citep{xiong2021nystromformer} \\
    \cellcolor{red!25}XLNet \citep{yang2019xlnet} & - & 88.2 & Table 16 in \citep{zaheer2020big} \\
    \cellcolor{yellow!25}BERT-base \citep{devlin2018bert} & - & 88.4 & Table 2 in \citep{xiong2021nystromformer} \\ 
    \cellcolor{yellow!25}BERT \citep{devlin2018bert} & - & 88.9 & Table 16 in \citep{zaheer2020big} \\
    Transformer (L24H1024) \citep{vaswani2017attention} & - & 90 & Table 2 in \citep{dai2020funnel} \\
    \cellcolor{blue!25}ROBERTABase / RoBERTa-base \citep{liu2019roberta} & - & 90.2 & Table 3 in \citep{dai2020funnel} / Table 16 in \citep{zaheer2020big} \\
    \cellcolor{red!25}XLNetLarge \citep{yang2019xlnet} & - & 90.8 & Table 3 in \citep{dai2020funnel} \\
    ELECTRALarge \citep{clark2020electra} & - & 90.8 & Table 3 in \citep{dai2020funnel} \\  
    \cellcolor{blue!25}ROBERTALarge \citep{liu2019roberta} & - & 90.9 & Table 3 in \citep{dai2020funnel} \\
    Funnel Transformer (B10-10-10H1024) \citep{dai2020funnel} & - & 90.9 & Table 3 in \citep{dai2020funnel} \\
    MPNetBase \citep{song2020mpnet} & - & 91.5 & Table 3 in \citep{dai2020funnel} \\
    BIGBIRD \citep{zaheer2020big} & - & \textbf{91.5} & Table 16 in \citep{zaheer2020big} \\
    \bottomrule
    \end{tabular}
\end{table}

\begin{table}[ht]
\caption{Results of empirical studies on \textbf{TriviaQA}, a popular question answering dataset. Bigger accuracy/F1 is better. Results are reported on three evaluation sets: \textbf{Dev}, \textbf{Test}, and \textbf{Test Verified}.}
\label{tab:full_triviaqa}
\begin{tabular}{p{0.5\textwidth}|l|l|p{0.31\textwidth}}
    \toprule
    \textbf{Model} & \textbf{Acc} $\uparrow$ & \textbf{F1} $\uparrow$ & \textbf{Sources} \\ \midrule
    On \textbf{Dev}: & & & \\
    \cellcolor{red!25}{T5-Base \citep{raffel2020exploring}} & 24.5 & - & Table 5 in \citep{fedus2022switch} \\
    \cellcolor{red!25}{T5-Large \citep{raffel2020exploring}} & 29.5 & - & Table 5 in \citep{fedus2022switch} \\
    \cellcolor{blue!25}{Switch-Base \citep{fedus2022switch}} & \textbf{30.7} & - & Table 5 in \citep{fedus2022switch} \\
    \cellcolor{blue!25}{Switch-Large \citep{fedus2022switch}} & \textbf{36.9} & - & Table 5 in \citep{fedus2022switch} \\
    \cellcolor{blue!25}{Switch-C \citep{fedus2022switch}} & \textbf{47.5} & - & Table 5 in \citep{du2022glam} \\
    \cellcolor{yellow!25}{GPT-3 Zero-shot \citep{brown2020language}} & 64.3 & - & Table 11 in \citep{du2022glam} \\
\bottomrule
\end{tabular}
\end{table}

\begin{table}[ht]
\centering
\caption{Results of empirical studies on \textbf{NQ (long answer)}, a popular question answering dataset. Bigger F1 score is better. Results are reported on two evaluation sets: \textbf{Dev} and \textbf{Test}.}
\label{tab:full_nq_long}
%
\end{table}

\begin{table}[ht]
\centering
\caption{Results of empirical studies on \textbf{NQ (short answer)}, a popular question answering dataset. Bigger F1 score is better. Results are reported on two evaluation sets: \textbf{Dev} and \textbf{Test}.}
\label{tab:full_nq_short}
\begin{tabular}{p{0.56\textwidth}|l|p{0.31\textwidth}}
    \toprule
    \textbf{Model} & \textbf{F1 $\uparrow$} & \textbf{Sources} \\
    \midrule
    On \textbf{Dev}: & & \\
    DecAtt \citep{parikh2016decomposable} + DocReader \citep{chen2017reading} & 31.4 & Table 2 in \citep{wang2021cluster}, Table 2 in \citep{zhang2021poolingformer} \\
    DocumentQA \citep{clark2017simple} & 35.7 & Table 2 in \citep{zhang2021poolingformer} \\
    DrQA \citep{chen2017reading} & 35.7 & Table 2 in \citep{wang2021cluster} \\
    BERT-base \citep{devlin2018bert} & 47.5 & Table 2 in \citep{ainslie2020etc} \\
    \cellcolor{yellow!25}{ETC-base \citep{ainslie2020etc}} & \textbf{52.2} & Table 2 in \citep{ainslie2020etc} \\
    BERT-large / BERT-large / BERT-joint \citep{alberti2019bert} & 52.7 & Table 2 in \citep{ainslie2020etc} / Table 2 in \citep{wang2021cluster} / Table 2 in \citep{zhang2021poolingformer} \\
\end{tabular}
    \midrule
    On \textbf{Test}: & & \\
    DecAtt \citep{parikh2016decomposable} + DocReader \citep{chen2017reading} & 31.5 & Table 2 in \citep{zhang2021poolingformer} \\ 
    DocumentQA \citep{clark2017simple} & 35.1 & Table 2 in \citep{zhang2021poolingformer} \\
    BERT-joint \citep{alberti2019bert} & 52.1 & Table 2 in \citep{zhang2021poolingformer} \\
    ETC (official) \citep{ainslie2020etc} & 57.86 & Table 5 in \citep{ainslie2020etc} \\
    BIGBIRD-ETC \citep{zaheer2020big} & 57.9 & Table 3 in \citep{zaheer2020big}, Table 2 in \citep{zhang2021poolingformer} \\
    Cluster-Former / Cluster-Former-ensemble \citep{wang2021cluster} & 60.9 & Table 3 in \citep{wang2021cluster} / Table 2 in \citep{zhang2021poolingformer} \\
    RikiNet-v2 / RikiNet-ensemble \citep{liu2020rikinet} & \textbf{61.3} & Table 3 in \citep{zaheer2020big} / Table 2 in \citep{zhang2021poolingformer} \\
    Poolingformer-esamble \citep{zhang2021poolingformer} & 61.6 & Table 2 in \citep{zhang2021poolingformer} \\
    ReflectionNet / ReflectionNet-ensemble \citep{wang2020no} & \textbf{64.1} & Table 3 in \citep{zaheer2020big} / Table 2 in \citep{zhang2021poolingformer} \\
    \bottomrule
\end{tabular}
\end{table}

\begin{table}[ht]
    \centering
    \caption{Results of empirical studies on \textbf{LRA-ListOps}. Bigger accuracy is better.}
    \label{tab:full_lra_listops}
    \scale[0.88]{\begin{tabular}{p{0.45\textwidth}|l|p{0.52\textwidth}}
    \toprule
    \textbf{Model} & \textbf{Acc} $\uparrow$ & \textbf{Sources} \\
    \midrule
    Local Attention / Local Attention / Local Attn.  \citep{tay2020long} & 15.82 & Table 1 in \citep{ma2021luna} / Table 10 in \citep{gu2022efficiently} / Table 1 in \citep{hasani2022liquid} \\
    Linear Trans. \citep{katharopoulos2020transformers} & 16.13 & Table 1 in \citep{ma2021luna}, Table 10 in \citep{gu2022efficiently}, Table 1 in \citep{hasani2022liquid} \\ 
    \cellcolor{red!25}Linformer \citep{wang2020linformer} & 16.13 & Table 1 in \citep{hasani2022liquid} \\ 
    Sparse Trans. / Sparse Trans. / Sparse Transformer \citep{child2019generating} & 17.07 & Table 1 in \citep{ma2021luna} / Table 10 in \citep{gu2022efficiently} / Table 1 in \citep{hasani2022liquid} \\ 
    \cellcolor{yellow!25}Performer in \citep{choromanski2020rethinking} & 18.01 & Table 1 in \citep{ma2021luna}, Table 2 in \citep{ma2023mega}, Table 10 in \citep{gu2022efficiently}, Table 1 in \citep{hasani2022liquid}, Table 3 in \citep{xiong2021nystromformer} \\
    \cellcolor{blue!25}Reformer \citep{kitaev2020reformer} & 19.05 & Table 3 in \citep{xiong2021nystromformer} \\ 
    \cellcolor{yellow!25}Performer in \citep{choromanski2020rethinking} & 32.78 & Table 1 in \citep{zhu2021long}\\
\bottomrule
\end{tabular}}
\end{table}

\begin{table}[ht]
    \centering
    \caption{Results of empirical studies on \textbf{LRA-Text}. Bigger accuracy is better.}
    \label{tab:full_lra_text}
    \scale[0.88]{\begin{tabular}{p{0.45\textwidth}|l|p{0.52\textwidth}}
    \toprule
    \textbf{Model} & \textbf{Acc} $\uparrow$ & \textbf{Sources} \\
    \midrule
    Local Attention / Local Attention / Local Attn. \citep{tay2020long} & 52.98 & Table 1 in \citep{ma2021luna} / Table 10 in \citep{gu2022efficiently} / Table 1 in \citep{hasani2022liquid} \\
    \cellcolor{red!25}Linformer \citep{wang2020linformer} & 53.94 & Table 1 in \citep{ma2021luna}, Table 2 in \citep{ma2023mega}, Table 10 in \citep{gu2022efficiently} \\
    \cellcolor{red!25}Linformer \citep{wang2020linformer} & 55.91 & Table 3 in \citep{xiong2021nystromformer} \\
    \cellcolor{yellow!25}Reformer \citep{kitaev2020reformer} & 56.1 & Table 1 in \citep{hasani2022liquid}, Table 10 in \citep{gu2022efficiently}, Table 2 in \citep{ma2023mega}, Table 1 in \citep{ma2021luna} \\
    \cellcolor{red!25}Linformer \citep{wang2020linformer} & 56.12 & Table 1 in \citep{zhu2021long} \\
    Sinkhorn Trans. \citep{tay2020sparse} & 61.2 & Table 1 in \citep{ma2021luna}, Table 10 in \citep{gu2022efficiently}, Table 1 in \citep{hasani2022liquid} \\
    Synthesizer \citep{tay2021synthesizer} & 61.68 & Table 1 in \citep{ma2021luna}, Table 10 in \citep{gu2022efficiently} \\
\bottomrule
\end{tabular}}
\end{table}

\begin{table}[ht]
    \centering
    \caption{Results of empirical studies on \textbf{LRA-Retrieval}. Bigger accuracy is better.}
    \label{tab:full_lra_retrieval}
    \scale[0.88]{\begin{tabular}{p{0.45\textwidth}|l|p{0.52\textwidth}}
    \toprule
    \textbf{Model} & \textbf{Acc} $\uparrow$ & \textbf{Sources} \\
    \midrule
    \cellcolor{red!25}Linformer \citep{wang2020linformer} & 52.27 & Table 1 in \citep{ma2021luna}, Table 2 in \citep{ma2023mega}, Table 10 in \citep{gu2022efficiently} \\
    Linear Trans. \citep{katharopoulos2020transformers} & 53.09 & Table 1 in \citep{ma2021luna}, Table 10 in \citep{gu2022efficiently}, Table 1 in \citep{hasani2022liquid} \\
    \cellcolor{red!25}Linformer \citep{wang2020linformer} & 53.09 & Table 1 in \citep{hasani2022liquid} \\
    Local Attention / Local Attention / Local Attn. \citep{tay2020long} & 53.39 & Table 1 in \citep{ma2021luna} / Table 10 in \citep{gu2022efficiently} / Table 1 in \citep{hasani2022liquid} \\
    \cellcolor{yellow!25}Reformer \citep{kitaev2020reformer} & 53.4 & Table 1 in \citep{hasani2022liquid}, Table 10 in \citep{gu2022efficiently}, Table 2 in \citep{ma2023mega}, Table 1 in \citep{ma2021luna} \\
    \cellcolor{blue!25}Performer in \citep{choromanski2020rethinking} & 53.82 & Table 1 in \citep{ma2021luna}, Table 2 in \citep{ma2023mega}, Table 10 in \citep{gu2022efficiently}, Table 1 in \citep{hasani2022liquid} \\
    Sinkhorn Trans. \citep{tay2020sparse} & 53.83 & Table 1 in \citep{ma2021luna}, Table 10 in \citep{gu2022efficiently}, Table 1 in \citep{hasani2022liquid} \\
\bottomrule
\end{tabular}}
\end{table}

\begin{table}[ht]
    \centering
    \caption{Results of empirical studies on \textbf{LRA-Image}. Bigger accuracy is better.}
    \label{tab:full_lra_image}
    \scale[0.88]{\begin{tabular}{p{0.45\textwidth}|l|p{0.52\textwidth}}
    \toprule
    \textbf{Model} & \textbf{Acc} $\uparrow$ & \textbf{Sources} \\
    \midrule
    \cellcolor{red!25}Performer in \citep{choromanski2020rethinking} & 37.07 & Table 3 in \citep{xiong2021nystromformer} \\
    \cellcolor{yellow!25}Linformer \citep{wang2020linformer} & 37.84 & Table 3 in \citep{xiong2021nystromformer} \\
    \cellcolor{blue!25}Reformer \citep{kitaev2020reformer} & 38.07 & Table 1 in \citep{hasani2022liquid}, Table 10 in \citep{gu2022efficiently}, Table 2 in \citep{ma2023mega}, Table 1 in \citep{ma2021luna} \\
    \cellcolor{green!25}Standard \citep{vaswani2017attention} & 38.2 & Table 3 in \citep{xiong2021nystromformer} \\
    \cellcolor{yellow!25}Linformer \citep{wang2020linformer} & 38.56 & Table 1 in \citep{ma2021luna}, Table 2 in \citep{ma2023mega}, Table 10 in \citep{gu2022efficiently} \\
    FNet \citep{lee2021fnet} & 38.67 & Table 10 in \citep{gu2022efficiently}, Table 1 in \citep{hasani2022liquid} \\ 
    BigBird \citep{zaheer2020big} & 40.83 & Table 1 in \citep{ma2021luna}, Table 2 in \citep{ma2023mega}, Table 10 in \citep{gu2022efficiently}, Table 1 in \citep{hasani2022liquid} \\
\bottomrule
\end{tabular}}
\end{table}

\begin{table}[ht]
    \centering
    \caption{Results of empirical studies on \textbf{LRA-Pathfinder}. Bigger accuracy is better.}
    \label{tab:full_lra_pathfinder}
    \scale[0.88]{\begin{tabular}{p{0.45\textwidth}|l|p{0.52\textwidth}}
    \toprule
    \textbf{Model} & \textbf{Acc} $\uparrow$ & \textbf{Sources} \\
    \midrule
    Local Attention / Local Attention / Local Attn. \citep{tay2020long} & 66.63 & Table 1 in \citep{ma2021luna} / Table 10 in \citep{gu2022efficiently} / Table 1 in \citep{hasani2022liquid} \\
    Sinkhorn Trans. \citep{tay2020sparse} & 67.45 & Table 1 in \citep{ma2021luna}, Table 10 in \citep{gu2022efficiently}, Table 1 in \citep{hasani2022liquid} \\
    \cellcolor{red!25}Linformer \citep{wang2020linformer} & 67.6 & Table 3 in \citep{xiong2021nystromformer} \\
    \cellcolor{yellow!25}Reformer \citep{kitaev2020reformer} & 68.5 & Table 1 in \citep{hasani2022liquid}, Table 10 in \citep{gu2022efficiently}, Table 2 in \citep{ma2023mega}, Table 1 in \citep{ma2021luna} \\
    H-Trans.-1D / H-Transformer-1D \citep{zhu2021h} & 68.78 & Table 2 in \citep{smith2022simplified} / Table 1 in \citep{hasani2022liquid} \\  
    \cellcolor{yellow!25}Reformer \citep{kitaev2020reformer} & 69.36 & Table 3 in \citep{xiong2021nystromformer} \\
    Synthesizer \citep{tay2021synthesizer} & 69.45 & Table 1 in \citep{ma2021luna}, Table 10 in \citep{gu2022efficiently} \\
\bottomrule
\end{tabular}}
\end{table}


\end{document}